\documentclass[twoside,11pt]{article}

\usepackage{blindtext}
\usepackage{soul}

\usepackage{jmlr2e}

\usepackage{graphicx}
\usepackage{booktabs}
\usepackage{multirow}
\usepackage{arydshln}
\usepackage{algorithm}     
\usepackage{algpseudocode} 
\usepackage{enumitem}
\usepackage{framed}
\usepackage[export]{adjustbox}
\usepackage[accsupp]{axessibility}  
\usepackage{hyperref}
\usepackage{subcaption}
\usepackage{cleveref}
\usepackage{colortbl}
\usepackage{xcolor}
\definecolor{darkblue}{rgb}{0.0, 0.0, 0.55}
\usepackage[acronym]{glossaries}
\usepackage[most]{tcolorbox}
\usepackage{orcidlink}
\algrenewcommand\algorithmicrequire{\textbf{Input:}}
\algrenewcommand\algorithmicensure{\textbf{Output:}}

\renewcommand{\footnotesize}{\scriptsize}

\glsdisablehyper
\makeglossaries
\newacronym{llms}{LLMs}{Large Language Models}
\newacronym{llm}{LLM}{Large Language Model}

\newcommand{\vanilla}{\texttt{Baseline}\xspace}
\newcommand{\expert}{\texttt{Expert-guided}\xspace}
\newcommand{\llmgen}{\texttt{LLM-guided}\xspace}
\newcommand{\novel}{\texttt{Novel-Mapping}\xspace}
\newcommand{\great}{GReaT\xspace}

\firstpageno{1}

\begin{document}

\title{On The Role of Prompt Construction In Enhancing Efficacy and Efficiency of LLM-Based Tabular Data Generation}

\author{
  \name Banooqa Banday\thanks{Equal contribution. Accepted to IEEE ICASSP 2025.} \email banooqa@txstate.edu \\
  \addr Texas State University 
  \AND
  \name Kowshik Thopalli\footnotemark[1] \email kowshik\_thopalli@llnl.gov \\
  \addr Lawrence Livermore National Laboratory 
  \AND
  \name Tanzima Z. Islam \email tanzima@txstate.edu \\
  \addr Texas State University 
  \AND
  \name Jayaraman J. Thiagarajan \email jjayaram@llnl.gov \\
  \addr Lawrence Livermore National Laboratory 
}

\editor{}

\maketitle
\begin{center}
    \textbf{}
\end{center}
\begin{abstract}
LLM-based data generation for real-world tabular data can be challenged by the lack of sufficient semantic context in feature names used to describe columns. We hypothesize that enriching prompts with domain-specific insights can improve both the quality and efficiency of data generation. To test this hypothesis, we explore three prompt construction protocols: \expert, \llmgen, and \novel. Through empirical studies with the recently proposed GReaT framework, we find that context-enriched prompts lead to significantly improved data generation quality and training efficiency. 
\end{abstract}


\section{Introduction}
\label{sec:intro}
\begin{figure*}[t]
    \centering    
    \includegraphics[width=0.85\textwidth]{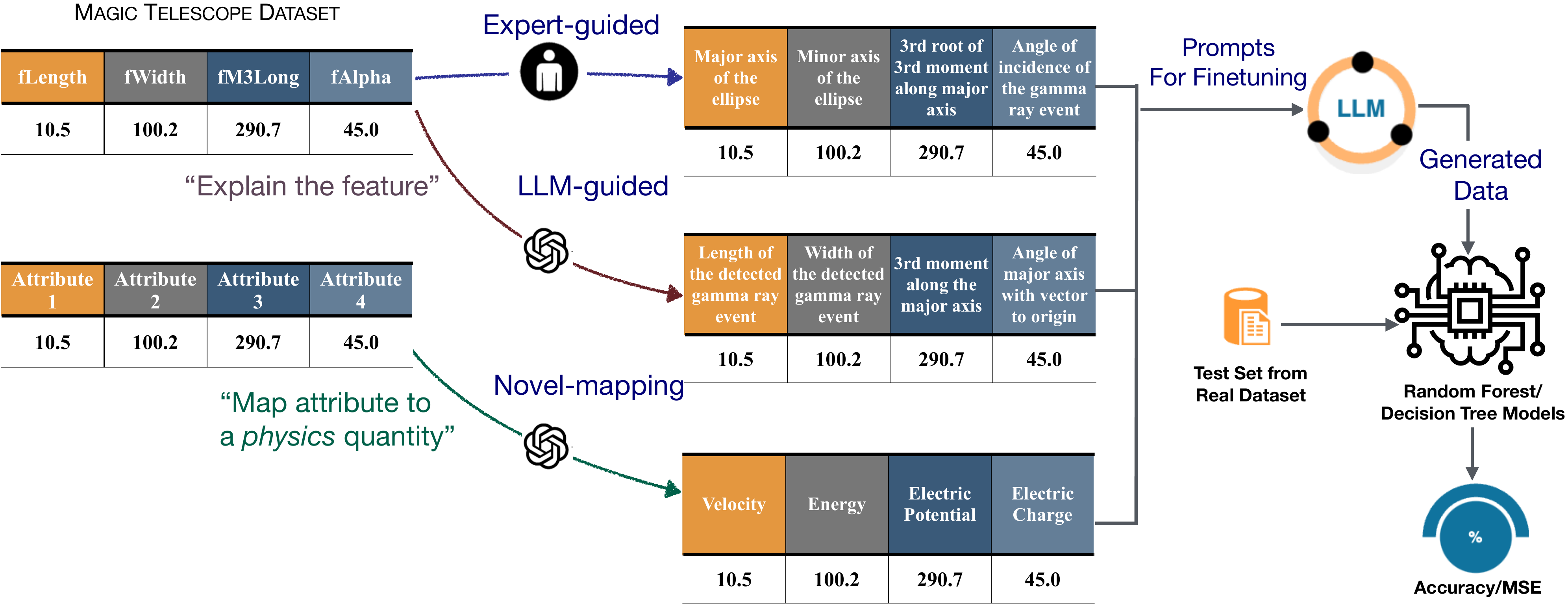}
    \vspace{0.005in}
    \caption{An overview of our approach for LLM-based tabular data generation. Our contributions include designing new prompt construction strategies and investigating their role in improving the quality of synthesized samples.}
    \label{fig:block_diagram}
    \vspace{-3truemm}
\end{figure*}

Generating realistic synthetic tabular data is a significant challenge with important applications in data augmentation~\citep{ding2024data,yuan2023large}, privacy preservation~\citep{rubaie_privacy_preservation}, and data imputation~\citep{li2024towards}. While a wide variety of approaches have been proposed, LLM-based tabular data generation has emerged a promising research direction. In this regard, Borisov \textit{et al.} recently proposed GReaT (Generation of Realistic Tabular data)~\citep{great}, which transforms tabular data into textual encodings (or prompts) and fine-tunes pre-trained LLMs to generate synthetic samples. 
More specifically, the text prompt in GReaT uses a subject-predicate-object schema, where the subject is simply the feature name. However, feature names in many real-world tabular datasets can be ambiguous, contain non-decipherable abbreviations or symbols, and even generic labels with no semantic context. In such cases, the text encoding used by GReaT can be insufficient to obtain high-fidelity synthetic samples, and can be highly sub-optimal in the case of generic labels (e.g., attribute A).  
In this paper, we hypothesize that enriching text prompts with domain-specific insights can significantly enhance an LLM's ability to synthesize high-quality tabular data. To validate this hypothesis, we propose three different prompt construction protocols (see Figure \ref{fig:block_diagram}): (i)~\expert, where domain experts provide detailed descriptors for feature names; (ii)~\llmgen, where external LLMs automatically generate feature descriptors for a given feature name and the dataset name; and (iii)~\novel which leverages an external LLM to map generic feature names, based on their value ranges to meaningful features from a novel domain (e.g., physics or life sciences). 

Through experiments on diverse datasets and two different LLMs, we demonstrate that our context-enriched prompting strategies consistently outperform the baseline of using raw feature names. The enhanced prompts not only improve the quality of the generated data, but also significantly boost training efficiency ($< 25\%$ of the epochs required by the baseline to achieve similar performance). Notably, the benefits persist even with parameter-efficient fine-tuning methods such as LoRA \citep{lora}. Finally, the~\novel protocol with no assumption on access to the actual feature names significantly outperforms the GReaT baseline even with access to the features.

\section{Background}
\label{sec:background}
\noindent\textbf{Problem Setup}:
Let $\mathcal{D} = \{(\mathbf{x}_i, y_i)\}_{i=1}^N$ denote a tabular dataset with $N$ samples, where $\mathbf{x}_i \in \mathcal{X}$ is a set of features in $\mathbb{R}^n$ and $y_i \in \mathcal{Y}$ is the corresponding target (categorical or continuous-valued). Without loss of generality, we refer to the names of the $n$ input features in the table as $\{c_1 \cdots c_n\}$. We aim to build a generative model that can synthesize realistic samples  
$\{(\mathbf{\tilde{x}}_j, \tilde{y_j})\}_{j}$, where $\mathbf{\tilde{x}}_j \in \mathcal{X}$ and $\tilde{y_j} \in \mathcal{Y}$. Given its broad applicability, this problem has gained significant attention, leading to the proposal of several solutions, with the most common being the extension of generative models from the vision literature to tabular datasets. Examples include TGAN~\citep{tgan}, CTGAN~\citep{ctgan}, TVAE~\citep{ishfaq2018tvae}, and even diffusion models~\citep{kotelnikov2023tabddpm}. However, recently, Borisov \textit{et al.} explored an alternative approach (GReaT) of fine-tuning pre-trained LLMs to build tabular data generators and demonstrated state-of-the-art capabilities. Since our study builds upon this approach, we now provide a brief overview of GReaT~\citep{great}.

\noindent\textbf{An Overview of GReaT}: In this approach, each $n-$dimensional input sample (a row in the input data matrix $\mathbf{X} \in \mathbb{R}^{T \times n}$) is first transformed into a textual encoding, and subsequently used as a prompt to query an LLM. 
This encoding strategy, which we refer to as~{\vanilla} encoding, constructs row-wise prompts by directly utilizing the feature names and adding the \texttt{is} qualifier to separate feature names and their corresponding values. For e.g., the encoding for the $i^{\text{th}}$ row of the input matrix $\mathbf{X}$ can be written as \textcolor{blue}{$``c_1 \texttt{ is } x^1_i, c_2 \texttt{ is } x^2_i, \cdots, c_n \texttt{ is } x^n_i"$}, where $x^k_i$ represents the value of the $k^{\text{th}}$ feature from the $i^{\text{th}}$ sample.~\great then fine-tunes pre-trained LLMs on these prompts using a next-token prediction objective. Once the model is fine-tuned, new samples can be unconditionally generated by post-processing the LLM's response for a test prompt that does not contain the feature values.

\section{Proposed Work}
\label{sec:approach}

While the~\vanilla encoding has been shown to lead to strong generation capabilities~\citep{great}, with real-world tabular datasets, the feature names are not always chosen to provide sufficient context for the observed values; for example, real-world datasets can contain ambiguous or generic feature names such as \textit{Attribute A}. Furthermore, it is common to use abbreviations or symbols that are not readily decipherable (e.g., \textit{fAlpha} in the magic telescope dataset) without sufficient expertise in the considered domain. In such cases, it can be challenging for an LLM to leverage useful priors from its pre-training, thereby impacting its generation performance. Consequently, the focus of this work is to study the impact of enriching the prompts with better clarity and specificity on an LLM's ability to generate high-quality tabular data. To this end, we explore three different prompt construction protocols: (i)~\expert: Expand feature names with domain-specific descriptors during prompt construction; (ii)~\llmgen: Leverage an LLM to provide additional description of the features based on their original names in the table; (iii)~\novel: Leverage an LLM to provide feature names given the original value ranges and the name of a field. This protocol is useful when the feature names are generic and do not contain any meaningful information to expand upon.

\subsection{Prompt Construction Protocols}

\noindent\textbf{(i)~\expert}: 
In this approach, we replace the feature names with user-provided descriptors detailing the semantic context of each column in the tabular dataset while retaining the \texttt{is} qualifier from~\vanilla encoding. Although this requires additional human effort,
our empirical study shows that this expanded context not only improves the efficacy of data generation but also provides significant gains in training efficiency, \textit{i.e.}, number of epochs for fine-tuning. 

\noindent\textbf{(ii)~\llmgen}: %
As an alternative to expert guidance, we also explore the feasibility of leveraging a pre-trained~\gls{llm} such as ChatGPT~\citep{openai2024chatgpt} to automate this process. More specifically, we query the ChatGPT API with the following prompt: \textcolor{blue}{``For a dataset named \textless name\textgreater, the given column names are  \textless list of column names\textgreater. You need to provide a short one-line description of each feature.''} The response from ChatGPT is then parsed and used in lieu of the original feature names $c_k$ during prompt construction for the fine-tuning step. Note that, this approach is applicable only when the feature names are at least partially specified (e.g., abbreviations or symbols). 

\noindent\textbf{(iii)~\novel}:
Finally, in realistic scenarios where the column names contain no useful information (e.g., Column A, Column B, $\cdots$), neither of the above two approaches will be applicable. In such a case, we propose the use of the \novel protocol, which will query an external LLM to generate a suitable feature name from an arbitrary domain (e.g., physics or life sciences); for example, one can use the query \textcolor{blue}{``I have a dataset that does not have meaningful names for features. Given the ranges of the columns are  \textless list of ranges\textgreater, suggest a term/phenomenon from  \textless field name\textgreater~that can take values in each of the given ranges. Rules are: (i) the terms/phenomenon should be from the same field, (ii) no two suggestions can be identical.''}. Note, the  \textless field name\textgreater~can be arbitrarily chosen as long as the feature names remain consistent with the LLM's prior knowledge (i.e., chosen from the same domain) and they have a similar range of feasible values ( \textless list of ranges\textgreater). Figure \ref{fig:block_diagram} illustrates an example with the Magic Telescope dataset, where we replace the generic attribute labels with quantities from physics.

\subsection{LLM Fine-tuning for Data Generation}
While GReaT~\citep{great}, by design, fine-tunes all LLM parameters, our study considers both regular fine-tuning as well as parameter-efficient fine-tuning (PEFT) based on LoRA~\citep{lora}. In a nutshell, LoRA is a technique for efficiently fine-tuning LLMs by restricting updates to a low-rank subspace of the model's gradient space, allowing significant parameter adaptation with minimal computational overhead.

\subsection{Implementation}
For this study, we used 
two LLMs, namely GPT-2~\citep{radford2019language} and DistilGPT-2~\citep{sanh2019distilbert} and build upon the publicly released~\great codebase~\footnote{https://github.com/kathrinse/be\_great}. For fine-tuning DistilGPT-2, we used the AdamW optimizer with learning rate $5e-5$ and trained for $400$ epochs. For GPT-2, we used LoRA with learning rate set to $5e-5$, $r = 16$ and $\alpha=32$. Our implementation utilizes the Transformers~\citep{huggingface_transformers} and PEFT~\citep{peft} libraries~\footnote{https://github.com/huggingface/}. 
\begin{table}[h]
\centering
\renewcommand{\arraystretch}{1.3}
\footnotesize
    \caption{Summary of datasets considered in this study.}
\resizebox{0.7\columnwidth}{!}{
\begin{tabular}{|c|c|c|}
        \hline
        \textbf{Dataset} & \textbf{Features} & \textbf{Targets} 
        \\
        \hline
        HELOC & &  Likelihood of loan repayment\\ 
        \citep{heloc} & \multirow{-2}{*}{23} & (classification)
        \\\hline
        Magic Gamma Telescope &  & Class label -- gamma ray or cosmic ray \\
        \citep{misc_magic_gamma_telescope_159} & \multirow{-2}{*}{10} & (classification) 
        \\ \hline
        California Housing &  & Median house value \\
        \citep{cal_housing} & \multirow{-2}{*}{8} & (regression)
        \\\hline
        Parkinson's Diagnosis& & Parkinson's score \\
        \citep{misc_parkinsons_telemonitoring_189}  & \multirow{-2}{*}{19}  & (regression)
        \\\hline
    \end{tabular}
}
\label{tab:datasets}
\end{table}


\section{Experimental Setup}
\label{sec:experimental_setup}

\noindent \textbf{Datasets.} Table~\ref{tab:datasets} summarizes the four tabular datasets considered in this work, comprising both classification and regression tasks. We provide detailed descriptions of the datasets in the appendix.

\noindent \textbf{Evaluation.} To assess the quality of the synthetic data generated with our prompting strategies, we test how well predictive models trained solely on this synthetic data perform on real test data. In prior work, this evaluation has been referred to as machine learning efficiency (MLE)~\citep{great}. To estimate MLE, we utilize two widely used ML models for tabular data -- random forests (RF) ~\citep{randomforest} and decision trees (DT)~\citep{decision_tree}. We train these models using the sklearn~\citep{scikit-learn} library and conduct hyper-parameter tuning through grid-search with 5-fold cross-validation. As evaluation metrics, we use the mean squared error and accuracy scores for regression and classification tasks respectively. MLE quantifies how well models trained purely on the synthetic data can generalize to real unseen data, thereby serving as an effective proxy for the quality of the generated samples.

\section{Results and Findings}
\label{sec:results}
In this section, we perform a comprehensive evaluation of the impact of the prompt construction methods on MLE. We present our findings below

\noindent \textcolor{darkblue}{\textbf{Finding 1:} \textit{Leveraging semantic context in prompts boosts LLM-based data generation}}.
\newline
\noindent Table~\ref{tab:mle-prompts-distilgpt2} shows the MLE scores achieved by models trained on synthetic data generated using the different prompting methods after fine-tuning all the parameters of DistilGPT-2. For the Magic Telescope dataset, the~\expert prompts boost accuracy by up to $3.3\%$ 
over the baseline and outperforms the~\llmgen prompts by $2.4\%$ 
for the RF model. 
\begin{table}[t]
\centering
\caption{Prediction performance of decision tree and random forest models on four datasets. ML models are trained on data generated by fine-tuning Distil-GPT2. Results demonstrate that 
enriching prompts with relevant semantic context 
yields a boost in performance.}
\resizebox{0.8\columnwidth}{!}{%
\begin{tabular}{c|c|cc}
\hline
\textbf{Dataset} & \textbf{Prompting} & \multicolumn{2}{c}{\textbf{Performance}}       \\
\cline{3-4}
 \textbf{(Metric)}& \textbf{Protocol} &\textbf{Decision Tree} & \textbf{Random Forest }\\
\hline \hline
      \multirow{3}{*}{\begin{tabular}[c]{@{}c@{}}Magic Telescope\\ (Accuracy)\end{tabular}} & \vanilla                          & 80.57         & 82.94         \\
                                                                                      & \expert                 &  \textbf{82.1 }         &  \textbf{86.25   }      \\
                                                                                      & \llmgen                             & 80.6          & 83.81         \\

                                                                                      \hline 
\multirow{3}{*}{\begin{tabular}[c]{@{}c@{}}HELOC \\ (Accuracy)\end{tabular}}          & \vanilla                       & 69.12         & 70.65         \\
                                                                                      & \expert                 & 69.22         &  \textbf{70.7 }         \\
                                                                                      & \llmgen                             &  \textbf{69.36}         & 70.27         \\
                                                                                      \hline 
\multirow{3}{*}{\begin{tabular}[c]{@{}c@{}}Parkinsons Diagnosis\\ (MSE)\end{tabular}}           & \vanilla                    & 11.15         & 10.2          \\
                                                                                      & \expert                  & 4.21          & 1.96          \\
                                                                                      & \llmgen                              &  \textbf{3.52}          &  \textbf{1.84}          \\
                                                                                      \hline 
\multirow{3}{*}{\begin{tabular}[c]{@{}c@{}}California Housing\\ (MSE)\end{tabular}}   & \vanilla                         & 0.5           & 0.35          \\
                                                                                      & \expert                  &  \textbf{0.46 }         &  \textbf{0.34}          \\
                                                                                      & \llmgen                             & 0.48          & \textbf{ 0.34}          \\
\hline
\end{tabular}
}
\label{tab:mle-prompts-distilgpt2}
\end{table}

\begin{figure}[h!]
    \centering
    \includegraphics[width=0.5\linewidth]{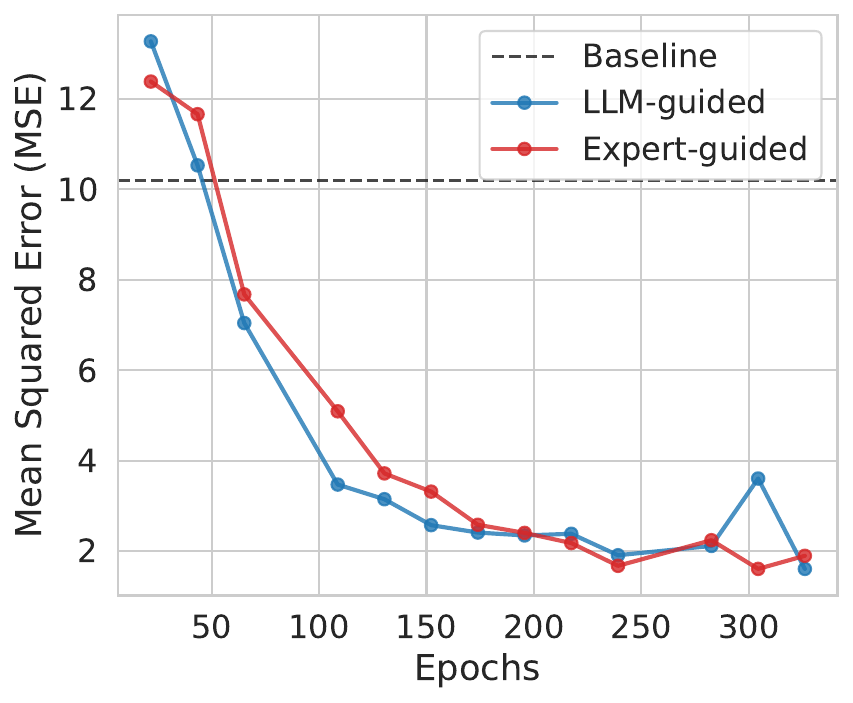}
    \caption{Enhanced prompt construction strategies lead to better computational efficiency.}
    \label{fig:efficiency}
\end{figure}

\noindent \textcolor{darkblue}{\textbf{Finding 2:} \textit{With better prompts, training efficiency comes for free}}.
\newline
\noindent Figure \ref{fig:efficiency} provides insights into the training dynamics when using the proposed prompting strategies on the Parkinson's diagnosis dataset. Strikingly, with both~\llmgen and~\expert prompts, the models surpass the MLE while requiring $<25\%$ of the \vanilla training epochs.

On the Parkinson's diagnosis dataset, the expert-guided and LLM-guided approaches 
reduce the MSE by $> 80\%$ compared to the~\vanilla encoding for both ML models. Enhanced prompts do not provide significant performance gains on the HELOC and California Housing datasets, which already contain non-ambiguous and readily interpretable feature names.

\begin{figure}[h]
    \centering
    \includegraphics[width=0.7\linewidth]{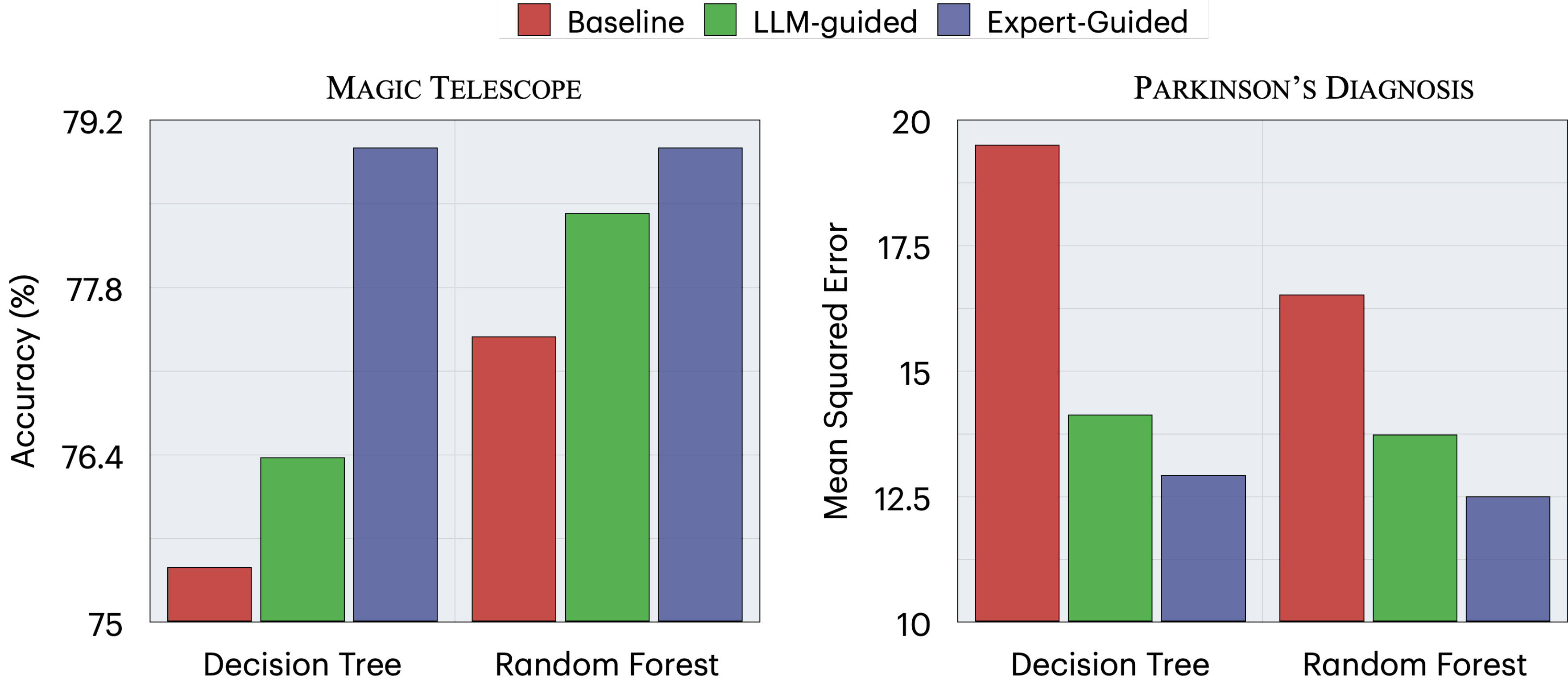}
    \caption{
    Performance of ML models trained on synthetic data, generated by fine-tuning GPT-2 with LoRA using various prompting methods, evaluated on the Magic Telescope and Parkinson's diagnosis datasets.
    }
    \label{fig:mle-prompts-gpt2-lora}
\end{figure}

\noindent \textcolor{darkblue}{\textbf{Finding 3:} \textit{Benefits continue to persist even with parameter-efficient fine-tuning}}.
\newline
\noindent Figure~\ref{fig:mle-prompts-gpt2-lora} presents the MLE scores achieved by models trained on synthetic data generated from the GPT-2 model fine-tuned with LoRA.

A striking observation is that the proposed prompting strategies continue to outperform~\vanilla encoding even with PEFT. For example, on the Parkinson's diagnosis dataset, the~\expert and~\llmgen prompts reduce the MSE by $33.7\%$ and $27.5\%$, respectively, compared to \vanilla prompting for the DT model. Furthermore, on the Magic Telescope dataset,~\expert prompts achieve a non-trivial accuracy boost of $3.5\%$. 

\begin{figure}[h]
    \centering
    \vspace{0.05in}
    \includegraphics[width=0.6\linewidth]{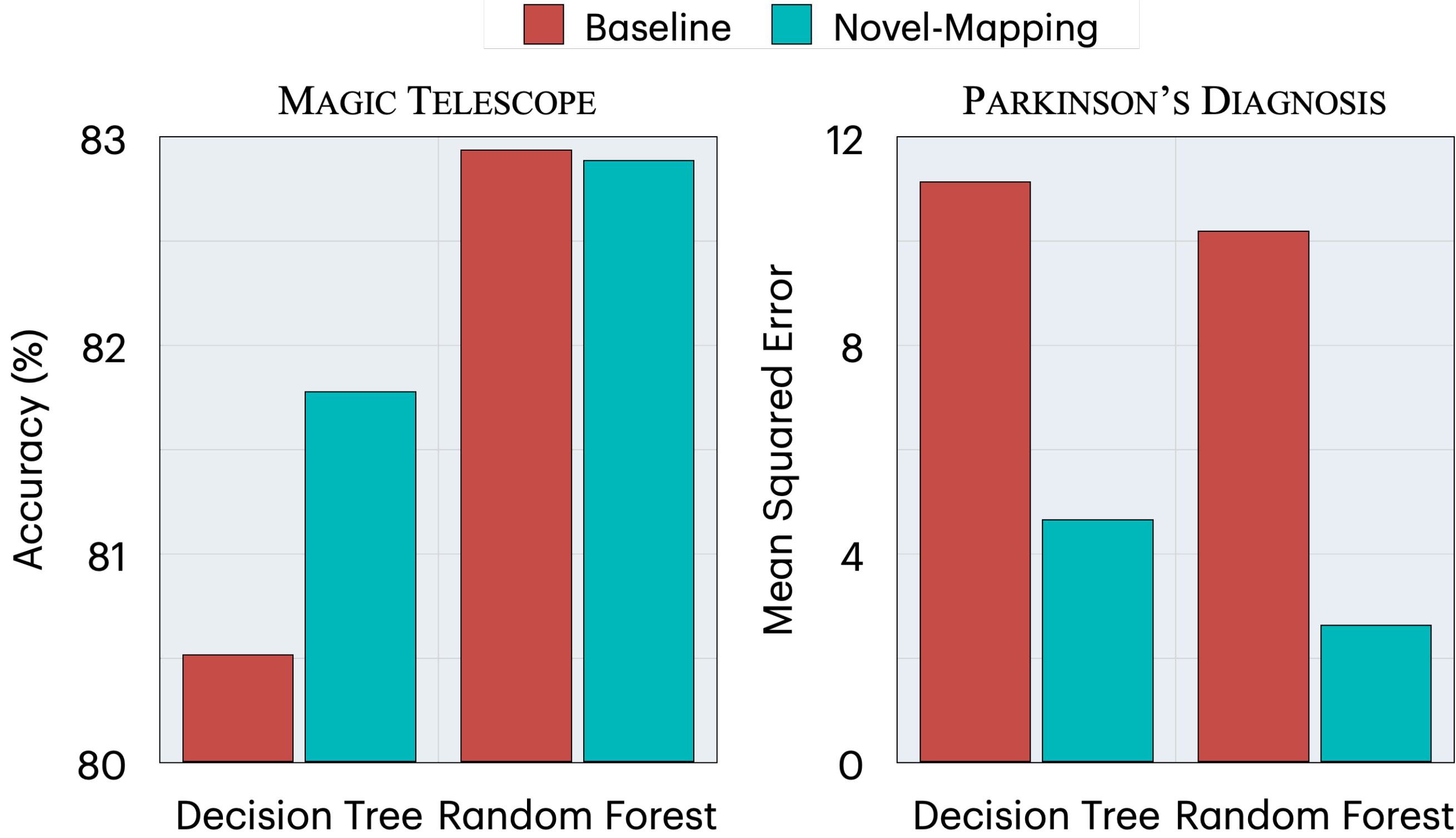}
    \caption{Mapping generic feature names to semantically meaningful descriptors from a novel domain provides non-trivial gains in performance.}
    \label{fig:novel}
\end{figure}

\noindent \textcolor{darkblue}{\textbf{Finding 4:} \textit{When no context is available},~\novel \textit{is highly effective}}.
\newline
\noindent In Figure~\ref{fig:novel}, we present the downstream prediction performance obtained with the~\vanilla and the~\novel strategies. Notably, when dealing with datasets containing non-decipherable names, mapping those feature names to meaningful names from another novel domain that is consistent with the priors of the pre-trained LLM provides significant benefits. For instance, in the case of the Magic Telescope dataset, we observe an accuracy improvement of $1.3\%$ points for the DT model, and similarly, for Parkinson's diagnosis, we observe substantial reductions of \textgreater$57\%$ in MSE for both models. While we establish the benefits of the~\novel strategy here compared to the~\vanilla with non-decipherable feature names, we note that this method will lead to even higher gains when the feature names are completely generic and lack relevant context. 

\section{Conclusions}
\label{sec:conclusions}
Our empirical results clearly evidence that when the feature names in tabular datasets do provide sufficient semantic context, the proposed prompting strategies can substantially enhance the quality of the generated samples. Furthermore, these strategies also exhibit improved computational efficiency. Interestingly, even \novel is a viable strategy in practice, particularly when the dataset contains only generic attribute descriptors.

\section{Limitations}
\label{sec:limitations}
We highlight some of the limitations of our approach which warrant further investigation. First, while we considered a diverse set of datasets, we only focus on four of them, and considering a more diverse range of datasets is required.  Second, we primarily assess the quality of the generated data using the Machine Learning Efficiency (MLE) metric, which need not capture all aspects of data quality. Incorporating additional evaluation metrics could provide a more comprehensive understanding of the generated data's properties. Finally, while we propose the \llmgen and \novel strategies to address the limitation of relying on human expertise in the \expert approach, further research is needed to validate their effectiveness across a wider range of scenarios.

\acks{This work was performed under the auspices of the U.S. Department of Energy by the Lawrence Livermore National Laboratory under Contract No. DE-AC52-07NA27344, Lawrence Livermore National Security, LLC. Supported by LDRD project 24-FS-002. LLNL-CONF-862086. This material is also based upon work supported by the U.S. Department of Energy, Office of Science under Award Number DE-SC0022843.}

\bibliography{main}
\newpage

\appendix
\section{Detailed Descriptions of the Datasets}
\subsection{HELOC (Home Equity Line Of Credit) } This has 10,459 samples and 23 features. The features in this dataset include detailed information about loan applicants, while the target variable indicates whether the applicants are likely to repay their loan within two years. This dataset is shared under CC license~\citep{heloc}.
\subsection{Magic Gamma Telescope} This dataset~\citep{misc_magic_gamma_telescope_159} comprises of ten features and one class label. The primary purpose of this dataset is to simulate the registration of gamma particles in the Cherenkov gamma telescope through imaging, thereby statistically differentiating between the particle showers caused by gamma rays (class g) and those caused by cosmic rays in the upper atmosphere (class h). The class column, indicating whether a sample belongs to class `h' or class `g', serves as the target variable. The dataset contains a total of 19,020 samples and is shared under CC BY 4.0 license.

\subsection{California Housing}  This is a dataset~\citep{cal_housing} based on census data from 1990 related to California and is shared under CC0 license. The features in this dataset include longitude and latitude representing specific areas, house age, number of rooms, total number of bedrooms, median household income, and number of people in the household. The target variable for this dataset is the median house value, which is a continuous value. The dataset comprises a total of 20,640 samples.

\subsection{Parkinsons Diagnosis} Parkinsons Diagnosis\citep{misc_parkinsons_telemonitoring_189} is a medical dataset consisting of voice measurements from individuals in the early stages of Parkinson's disease, recorded in their homes. This dataset includes 19 features: five features with the prefix "Jitter" related to frequency measurements, six features with the prefix "Shimmer" related to amplitude measurements, and two features corresponding to the ratio of noise to tonal components in the voice samples. The remaining features provide information on the patient's age, gender, and the time in days since enrollment. The target variable is the clinician's score, which is based on the other features, making this a regression dataset. The dataset contains a total of 5,875 samples.

We created real-train and real-test splits from all the datasets using a $90$-$10$ split. While the LLM was fine-tuned on the real-train split, for measuring MLE we train the ML models on the synthetic data and test on the real-test split.

\end{document}